%% file: root.tex
\documentclass[letterpaper, 10 pt, conference]{ieeeconf}  

\IEEEoverridecommandlockouts                              

\usepackage{color}
\usepackage{graphicx}
\usepackage{xcolor}
\usepackage{amsmath,amssymb,amsfonts,mathrsfs}
\usepackage[colorlinks=true, allcolors=blue]{hyperref}
\usepackage[caption=false,font=footnotesize,labelfont=sf,textfont=sf]{subfig}
\usepackage{stfloats}
\usepackage{algorithm}
\usepackage{algorithmic}
\usepackage{multirow}
\usepackage{pgf}
\usepackage[absolute,overlay]{textpos}
\usepackage{tikz}
\usepackage{xspace}
\usepackage{siunitx}
\newcommand{\EFAs}{EFAs\xspace}

\newcommand{\domv}{\ensuremath{\hat{V}}\xspace}

\newtheorem{definition}{Definition}

\usetikzlibrary{arrows,automata,calc,petri,shapes,positioning}
\usetikzlibrary{shapes.multipart}
\usetikzlibrary{shapes.misc}
\tikzset{automatonFig/.style={->,>=stealth',shorten >=1pt,auto,node distance=1.7cm,
                    thick,initial text={}}}
\tikzset{forbidding cross/.style={cross out,draw,-,
         minimum size=0.8cm,thin}}

\title{\LARGE \bf
Hazard Analysis of Collaborative Automation Systems:\\ A Two-layer Approach based on Supervisory Control and Simulation
}

\author{Tom P. Huck$^{1}$, Yuvaraj Selvaraj$^{2}$, Constantin Cronrath$^{2}$, Christoph Ledermann$^{1}$,\\Martin Fabian$^{2}$~\!\IEEEmembership{Senior Member,~IEEE}, Bengt Lennartson$^{2}$~\!\IEEEmembership{Fellow,~IEEE},\\ and Torsten Kröger~\IEEEmembership{Senior Member,~IEEE}
\thanks{*This work was supported by the German Federal Ministry for Economic Affairs and Climate Action under the project ``SDM4FZI'' (\url{www.sdm4fzi.de}), by FFI, VINNOVA under grant number 2017-05519 and under the ITEA3 ``AITOC'' project, and by the Wallenberg AI, Autonomous Systems and Software program (WASP) funded by the Knut and Alice Wallenberg Foundation. The support is gratefully acknowledged.}
\thanks{$^{1}$Intelligent Process Automation and Robotics Lab, Institute of Anthropomatics and Robotics (IAR-IPR), Karlsruhe Institute of Technology, Germany.
{\tt\small tom.huck@kit.edu}}%
\thanks{$^{2}$Division of Systems and Control, Department of Electrical Engineering, Chalmers University of Technology, Gothenburg, Sweden.}%
}

\begin{document}

\begin{textblock*}{12cm}(5cm,0.5cm) 
	\centering
	This work has been submitted to the IEEE for possible publication. Copyright may be transferred without notice, after which this version may no longer be accessible.
\end{textblock*}

\maketitle
\thispagestyle{empty}
\pagestyle{empty}

\begin{abstract}
Safety critical systems are typically subjected to hazard analysis before commissioning to identify and analyse potentially hazardous system states that may arise during operation.
Currently, hazard analysis is mainly based on human reasoning, past experiences, and simple tools such as checklists and spreadsheets. 
Increasing system complexity makes such approaches decreasingly suitable.
Furthermore, testing-based hazard analysis is often not suitable due to high costs or dangers of physical faults. 
A remedy for this are model-based hazard analysis methods, which either rely on formal models or on simulation models, each with their own benefits and drawbacks. 
This paper proposes a two-layer approach that combines the benefits of exhaustive analysis using formal methods with detailed analysis using simulation. 
Unsafe behaviours that lead to unsafe states are first synthesised from a formal model of the system using Supervisory Control Theory.
The result is then input to the simulation where detailed analyses using domain-specific risk metrics are performed. 
Though the presented approach is generally applicable, this paper demonstrates the benefits of the approach on an industrial human-robot collaboration system. 
\end{abstract}
\section{INTRODUCTION}
\label{sec:introduction}
Automation systems frequently interact with humans to achieve shared goals. Such collaborative systems are often safety-critical, as technical or human errors, design flaws, or combinations thereof can endanger humans. Normative requirements state that safety critical systems must be subjected to a \textit{hazard analysis} before commissioning \cite{STD_ISO2011,STD_IEC61508}. This is a design-time procedure to identify potentially unsafe system states that may arise during system operation. While the complexity of collaborative systems is increasing, development of hazard analysis methods lags behind. Hazard analyses are still often based on human reasoning and simple tools like
checklists \cite{Hornung2021}. However, this does not scale well with increasing system complexity \cite{RA_Leveson2012}. Since testing under real-world conditions is often infeasible (e.g., unavailable prototypes, dangers associated with real-world tests, or too many required test cases), \textit{model-based hazard analysis methods}
support human reasoning. Most model-based hazard analyses proposed in recent years are based either on \textit{formal models} (e.g. \cite{Askarpour2016,RA_Askarpour2021,RA_Rathmair2021}) or \textit{simulation models} (e.g. \cite{RA_Araiza2016,RA_Bobka2016a,RA_Huck2021}). The differences are in the level of abstraction and the achievable degree of coverage~\cite{Dill1998}. Formal models are based on abstract mathematical-logical representations (e.g., automata, transition systems) often allowing exhaustive checks of the entire model state space that result in guaranteed safety properties. However, physical safety aspects (e.g., collision geometries or forces), are frequently abstracted away to keep the analysis tractable.
In contrast, simulation models use more detailed representations (3D-models, physics engines, etc.). While they are still abstractions, the increased level of detail enables more accurate analyses than formal models. Nevertheless, exhaustive analyses with formal guarantees are rarely feasible because detailed simulation models are computationally expensive. In short: there is a fundamental trade-off between \textit{completeness} and \textit{accuracy} of hazard analyses.

We address this trade-off by proposing a two-layer approach that analyses the system on two abstraction levels. We use formal models to describe the system on an abstract level and infer from that potential hazards, which are then examined closer in simulation. We thereby combine the benefits from exhaustive analysis using formal methods and a detailed analysis using simulation. 

Behaviours that lead to potentially unsafe states are synthesised from the formal model using \emph{Supervisory Control Theory} (SCT)~\cite{ramadge1989control}. Although SCT is a formal approach to synthesise feedback controllers for \emph{discrete event systems} (DES), in this paper we re-purpose synthesis algorithms from SCT to identify unsafe behaviours that lead to hazardous states. We aim to compensate limited detail of the formal model by modelling the system conservatively, so that the synthesis yields an over-approximate set of unsafe behaviours. The potentially unsafe behaviours thus synthesised are then used as input for the simulation where they are analysed in more detail.

We demonstrate and evaluate our approach using examples from Human-Robot Collaboration (HRC). Our experiments show that even with relatively simple formal models, the two-layer approach finds significantly more hazards than simulation-only methods. Our experiments also highlight the trade-off between modelling abstractions and accuracy.

While
this paper focuses on HRC systems, 
our approach is transferable, as its foundational framework, SCT, is applicable to any system that can be modelled as a DES.

\section{RELATED WORK}
Various methods to analyse safety-critical systems are known. For industrial machinery, analyses are based mostly on human reasoning and expert knowledge, supported by simple tools (e.g., checklists) \cite{Hornung2021,STD_ISO2011}. The chemical industry uses Hazard and Operability Analysis (HAZOP), an analysis method combining human reasoning with system diagrams and guide words for structured analysis~\cite{STD_IEC61882}. HAZOP-UML is an extension of HAZOP methodology that uses UML-methods as system description~\cite{Guiochet2016}. A similar method is Systems-Theoretic Process Analysis (STPA) \cite{RA_Leveson2012} which also relies on human reasoning, system diagrams, and guide words, but uses control structure diagrams instead of flow diagrams. To support human analyses, there are also expert systems that encode domain-specific knowledge in rule-bases that map sets of system properties to corresponding hazards and suggested safety measures \cite{RA_Awad2017}.

Formal methods such as model checking \cite{BaiKat:08,MISC_Clarke2018} express systems and safety specifications in strictly formalised representations (e.g., automata and linear temporal logic) so that systems can be automatically checked against their specifications. While model checking has traditionally been applied to check software and hardware designs, it can also be applied to identify hazards in cyber-physical systems like robots \cite{Askarpour2016,Askarpour2017,RA_Askarpour2020,RA_Rathmair2021}. 

When systems are highly complex or contain black-box components, formal methods are not always tractable. In these cases, simulation-based testing is a suitable alternative \cite{Corso2020Survey}. Simulation-based analyses are especially common in the domain of autonomous vehicles \cite{Norden2019,Ding2020,Chance2019}, but are also found in robotics \cite{RA_Araiza2016,RA_Bobka2016a,RA_Huck2021} and aerospace \cite{Lee2015}.

Combinations of formal methods and simulations have been proposed in~\cite{Webster2020} and~\cite{Askarpour2020}, where simulations are combined with the model checkers PRISM and ZOT, respectively. However, these model checkers are only suitable for iterative procedures, where the model checker returns a single example sequence leading to an unsafe state (called an error trace). The human user then has to inspect a simulation of that sequence, fix the underlying safety flaw, and re-run the model checker. Our SCT-based approach differentiates itself from these works by automatically generating complete \textit{sets} of unsafe behaviours rather than individual examples.

\section{TWO-LAYER HAZARD ANALYSIS} \label{sec:method}
We now introduce the proposed hazard analysis approach. Sec.~\ref{subseq:DES} addresses the layer of DES, while Sec.~\ref{subseq:simulation} discusses the simulation layer.
\subsection{First Layer: Synthesis of Unsafe Behaviours}\label{subseq:DES}
\input{methodA_YS.tex}

\subsection{Second Layer: Simulation of Unsafe Behaviours}
\label{subseq:simulation}
While the previous step already gives insights into potential hazards, certain safety-critical aspects are abstracted away in the DES model. Often, a definitive judgement whether a sequence is safe or unsafe is only possible from a more detailed analysis. We therefore perform a second analysis step in simulation.
Note that this step is naturally domain-specific as each domain uses their own simulators. Thus, we only outline the general idea.

We create a simulation model where each system component, previously represented by an EFA, is now represented by a corresponding component in the simulation and each event from an EFA is associated with a certain activity of the respective component (e.g. an event from an EFA model of a human may correspond to a certain activity of a digital human model in the simulation). This correspondence allows for recreating the previously synthesised behaviours in simulation. However, directly recreating event sequences in a strictly sequential manner only adds limited value, since it does not consider the temporal nature of interactions between components in sufficient detail (e.g., whether a safety function responds quickly enough to avoid an unsafe state). 
The following example illustrates the necessity of the second simulation layer. Consider the following sequence of events:
\begin{gather*}
    (\mathit{activateRobot}, \mathit{approachRobot},\\ \mathit{robotStops}, \mathit{enterWorkspace})
\end{gather*}
If we simulate this strictly sequential, event by event, it will seem like the robot stops before the human enters the workspace. However, this may actually not be the case considering the robot's stopping time. Thus, we divide the events into a set of \textit{proactive} and \textit{reactive} events, where proactive events can occur without external stimuli, while reactive events are consequences of proactive events. We then extract from the synthesised behaviours only those events that are proactive and use them as input to control the simulation, while the reactive events are emergent reactions arising within the simulation.
In the example above, the proactive events are $(\mathit{activateRobot}, \mathit{approachRobot}, \mathit{enterWorkspace})$. These are used as inputs to drive the simulation. The reactive event ($\mathit{robotStops}$) is not prescribed as a simulation input, but emerges internally in the simulation as a consequence of the behaviour encoded in the robot's simulation model. Assuming that this model is accurate, we can thus determine whether the robot responds safely considering its actual behaviour, not the abstract behaviour of the formal model.

\input{automatonFig_combined}
\begin{figure}[ht!]
    \centering
    \includegraphics[width=1\columnwidth]{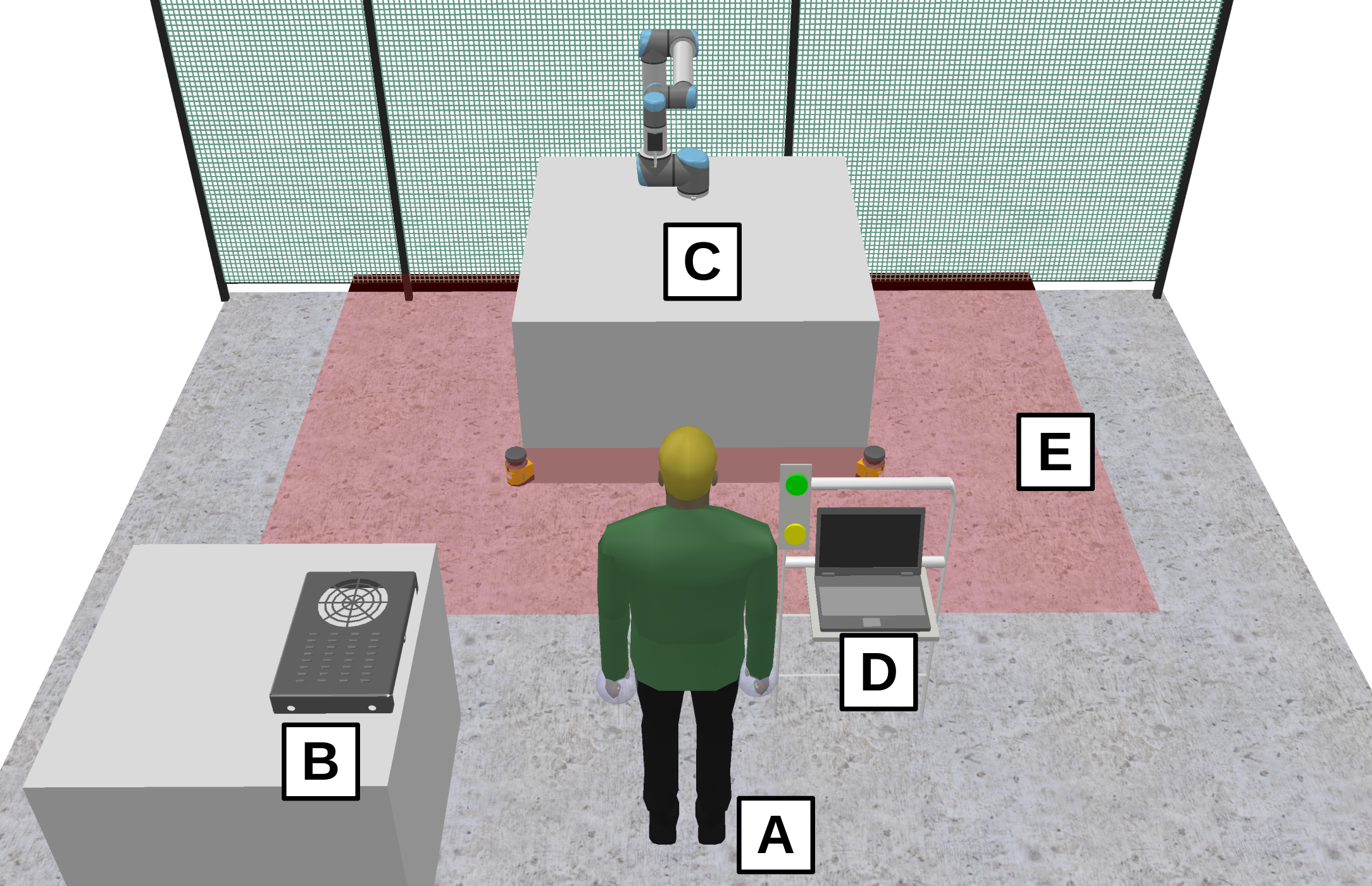}
    \caption{HRC system from the application example (A: centre area, B: parts storage, C: robot station, D: control panel, E: laser scanner zone)}
    \label{fig:scenario1}
\end{figure}

\section{APPLICATION EXAMPLE}
\label{sec:example}
In the following, we present an example to illustrate our approach. Models and code are available on GitHub\footnote{https://github.com/Huck-KIT/ICRA2023}. Consider a HRC workstation with formal models corresponding to Fig.~\ref{fig:example} and a simulation model as depicted in Fig.~\ref{fig:scenario1}. The collaborative task in this example is as follows: the human worker starts in the centre area (Fig.~\ref{fig:scenario1}, A), retrieves a part from the table (B), places it in front of the robot (C), then walks back to the centre area (A), and activates the robot at the control panel (D). The robot performs a procedure on the part until the worker stops it through the control panel. The worker then retrieves the part and places it back on the table. As a safety measure, the area around the robot is monitored by a laser scanner (E, red area). A safety stop of the robot is triggered when the worker enters the detection area. To inspect the ongoing procedure from close distance without triggering a safety stop, the worker can override this safety function through the control panel. We now want to identify possible hazards in this HRC system. To that end, we model the system through EFAs, define a safety specification, and perform a supervisor synthesis to find unsafe behaviours.

The models are shown in Fig \ref{fig:example}, with their locations (circles), events (arrows with black labels), guards (blue), and actions (red). For the the human $\mathcal{H}$, we introduce the following events: transiting between area A and B ($t_1$) and between A and C ($t_2$), as well as picking up/putting down the part at the storage table (up: $u_S$, down: $d_S$) and at the robot station (up: $u_R$ down: $d_R$), respectively. Also, the human can press several buttons to stop or activate the robot, which are modelled as events $b_0$ (stop), $b_1$ (start in normal mode), and $b_2$ (start in safety override mode). We introduce the variable $P\in\{0,1,2\}$ to track the part (0: at storage, 1: in worker's hands, 2: at robot station) and the variables $W\in\{0,1\}$ and $S\in\{0,1\}$ to track if the human currently occupies the shared human-robot workspace ($W$) and the laser scanner zone ($S$), respectively (0: not occupied, 1: occupied). Observe that some events require guard statements to be fulfilled before the transition is enabled. (e.g., to put down a workpiece at the robot station $d_R$, the guard is $P=1$, because the worker must first be in possession of the part to put it down). For the robot $\mathcal{R}$, a location $q_{\mathcal{R}}$ is introduced for each operation mode ($q_{\mathcal{R}0}$: idle, $q_{\mathcal{R}1}$: working, $q_{\mathcal{R}2}$: working in safety override mode). Pressing the buttons ($b_0,b_1,b_2$) causes the robot to change its operation mode (note that $b_0,b_1,b_2$ are shared events appearing both in $\mathcal{H}$ and $\mathcal{R}$). Additionally, the robot has an event $safety\ stop$ which is enabled if the robot is running in normal operation mode and the laser scanner is occupied (i.e., $S=1$). Note that $safety\ stop$ is not available in safety override mode (i.e., in $q_{\mathcal{R}2}$). The variable $R\in\{0,1\}$ tracks if the robot is currently idle (0) or active (1).
Note that in our models, the EFAs omit any information about the \textit{duration} of events. For instance, in $\mathcal{R}$, it is not determined whether the safety stop is executed immediately as the human enters the detection zone, or if the robot requires some stopping time. It is only stated that $S=1$ is a guard (i.e., a \textit{precondition}) for the safety stop. This is a deliberate measure to compensate for loss of accuracy due to modelling abstractions: by leaving the timing open, we force the synthesis to consider all possible interleavings of events when searching for unsafe sequences. Whether these sequences are indeed possible under the actual timing behaviour of the system is determined in simulation, the second analysis step (cf. our discussion in Sec. \ref{subseq:simulation}).

The safety specification ${\mathcal{SP}}$ simply has two locations $q_{\mathcal{SP}0}$ (safe) and $q_{\mathcal{SP}1}$ (unsafe). The unsafe location represents a collision. Observe that the unsafe location is \textit{marked}, but only reachable through event $c$ (collision), which has the guard statement $R=1\land W=1$. Thus, we consider a state to be a collision if the human is in the collaborative workspace and the robot is running at the same time (again, this is a conservative over-approximation to compensate for abstraction). We then perform a supervisor synthesis with respect to the plant $\mathcal{S}\parallel\mathcal{R}$ and the specification $\mathcal{SP}$. The resulting supervisor contains 90 states and 182 transitions. 

In the simulation step of our analysis, we consider the events related to the human as proactive, and those related to the robot as reactive (compare Sec. \ref{subseq:simulation}). We thus extract from the supervisor all human behaviours that lead to an unsafe state within a fixed time horizon of ten events, yielding 22 sequences (note that sequences in the supervisor can be arbitrarily long, as they may contain loops. To limit simulation time, we only extract sequences up to a fixed maximum length). One example for such an unsafe behaviour is the sequence $(b_2, r, t_1, u_s, r, t_1, t_2, d_R)$, where the human activates the robot in safety override mode, transits to the storage table, retrieves a part, transits back, then transits to the robot and places the part on the table, with the robot already running. With the robot in safety override mode, the event $safety\ stop$ is not enabled, thus leading to an unsafe state. We then simulate the extracted unsafe human behaviours to assess how hazardous they actually are when performed in conjunction with a more detailed and accurate model of the robot system. To quantify the level of danger without requiring a human user to inspect each simulation run, we compute a domain-specific \textit{risk metric} $r$:
\begin{equation}
	r = \begin{cases} 0 & \text{case (a):}\ v_R < v_{crit} \\
		\text{e}^{-d_{HR}} & \text{case (b):}\ v_R \geq  v_{crit};\ d_{HR}>0\\
		\frac{F_{\mathrm{c}}}{F_{\mathrm{max}}}+1 & \text{case (c):}\ v_R \geq  v_{crit};\ d_{HR}=0~.
	\end{cases}
	\label{eq:RiskMetric}
\end{equation}
where $d_{HR}$ is the human-robot distance, $v_R$ the robot speed, and $v_{crit}$ a speed threshold (here: \SI{250}{mm/s}). $F_{c}$ is the estimated human-robot collision force, and $F_{max}$ is a collision force limit according to \cite{STD_ISOTS15066}. The value of $r$ is calculated in each simulation time step, and the maximum value is recorded for each simulated sequence.

\section{EXPERIMENTS}
\label{sec:experiments}

\subsection{Experimental Setup}
\begin{figure}[b]
    \centering
    \includegraphics[width=1\columnwidth]{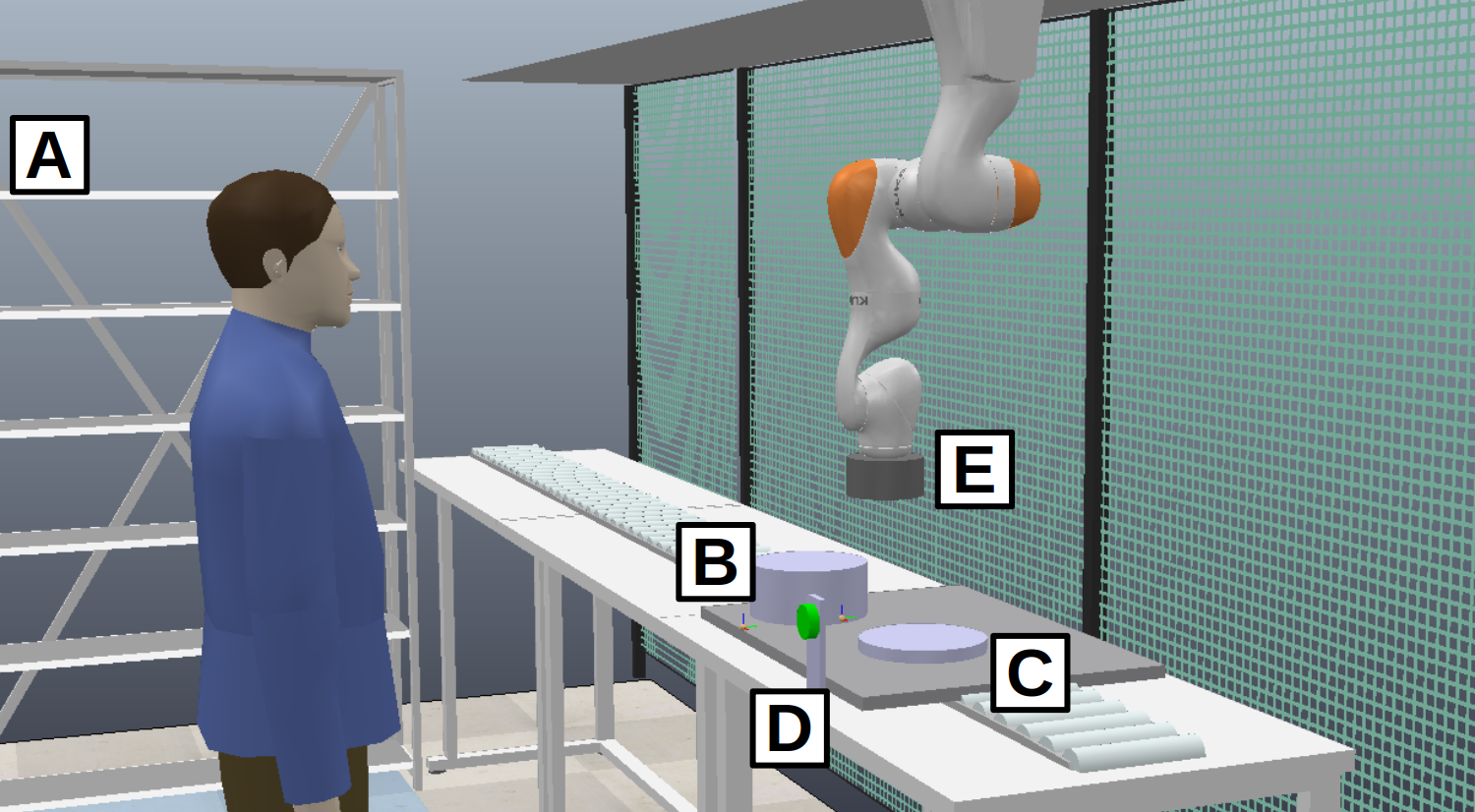}
    \caption{Second HRC scenario from the experiments. Here, the workflow is as follows: the worker retrieves parts from a shelf (A), inserts them into a housing and activates the robot with a button (D) which then inserts a gearwheel (E) into the housing. Meanwhile, the worker inserts a part into the cover (C) and finally mounts the cover onto the housing. Potential hazards consist in the hand being crushed between gearwheel and housing, and the head colliding with the robot's elbow joint. (further details on on GitHub and in the accompanying video).}
    \label{fig:scenario2}
\end{figure}

We validate our approach by conducting experiments in two test scenarios, including the example presented above (referred to as scenario A), as well as a second scenario with a more complex collaborative workflow where multiple parts of a workpiece are assembled collaboratively (scenario B), see Fig. \ref{fig:scenario2}. For brevity, we do not describe scenario B in detail. Details of both scenarios are shown GitHub$^{1}$. For testing purposes, each of the HRC systems is deliberately designed to contain some safety-critical flaws. For instance, in scenario A (see Sec. \ref{sec:example}), there are two flaws: first, there is a certain delay between the time the human enters the safety zone and the stop of the robot, leading to a possible collision hazard if the human approaches sufficiently fast; second, the safety override button allows the human can deactivate the safeguard, also leading to a collision hazard (although such an override button is unlikely to be found in a real robot system, we use it to introduce hazards for test purposes). We deploy our analysis described in Section \ref{sec:method} to find the sequences of events which lead to these unsafe states, using the tools \textit{Supremica} \cite{supremica2017} for EFA modelling and supervisor synthesis, and \textit{CoppeliaSim} \cite{CoppeliaSim2013} for simulation.

Performance criteria are as follows:
\begin{itemize}
    \item $N$: The number of sequences found that lead to a collision state.
    \item $r_{mean}$: The mean risk value of the unsafe sequences.
\end{itemize}
For comparison, we also deploy two further approaches. In contrast to the approach above, these approaches search for unsafe event sequences \textit{directly in simulation}, that is, without relying on a formal model as a first analysis step. In particular, we deploy random sampling, where events are sampled from a uniform distribution, and Monte Carlo Tree Search (MCTS). MCTS iteratively samples events, executes them in the simulator, and receives the resulting risk value (see (\ref{eq:RiskMetric})) as a reward. By attempting to maximise the reward (and thus, the risk that is associated with a sequence), MCTS finds sequences leading to unsafe states. This approach has been introduced in our earlier work, and we refer to \cite{RA_Huck2021} for detailed explanations.

\subsection{Test Runs and Results}
\begin{figure}
\centering
\begin{minipage}{.5\columnwidth}
  \centering
  \includegraphics[width=1\columnwidth]{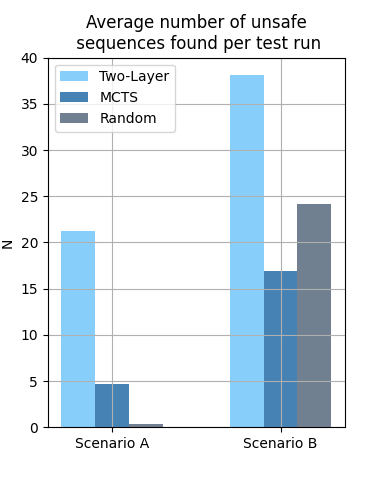}
\end{minipage}%
\begin{minipage}{.5\columnwidth}
    \centering
    \includegraphics[width=1\columnwidth]{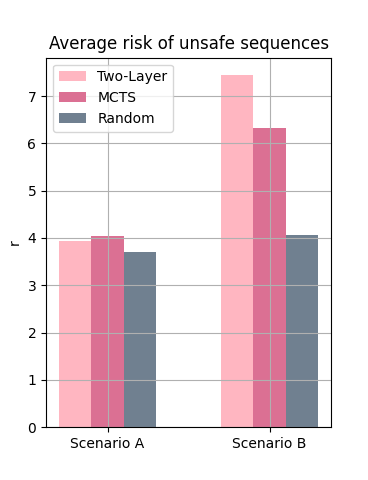}
\end{minipage}
\caption{Results from the test runs. Left: number of unsafe sequences found, right: average risk of unsafe sequences.}
\label{fig:results}
\end{figure}

Test runs for each of the three approaches (Two-layer, MCTS, Random) are executed with a maximum computational budget of 500 simulation runs, where each simulation run consists of up to 12 (10) events in scenario A (B). Since the computation time for synthesising the supervisor ($\leq$ \SI{1}{s}) is negligible compared to the simulation time (approx. \SI{30}{min} for 500 sequences), it is not accounted for in the budget.
To limit the influence of statistical outliers due to randomised features (e.g., randomly sampled motion parameters), each test run is repeated ten times with different random seeds. Fig. \ref{fig:results} shows results averaged over the test runs for each approach.

In both scenarios, the two-layer approach found significantly more unsafe sequences than the simulation-only approaches. Between the simulation-only approaches, MCTS performed better in scenario A, where unsafe sequences are relatively rare, while random sampling performed better in scenario B, where unsafe sequences are comparatively more frequent. In terms of risk, the two-layer approach on average achieved higher risk metrics than the simulation-only approaches in Scenario B, while average risks in Scenario A are relatively similar (a possible reason is that the scenario A is less diverse in terms of potential collision scenarios, which is reflected by less variation in the achievable risk).

We also investigated disagreements between the analyses on the two layers. In particular, we were interested in behaviours which falsely appear to be safe in the formal model, although they were found to be hazardous when analysed in simulation. We call these occurrences ``missed alarms", as the formal model is unable to bring the respective hazard to awareness. We also investigate the occurrence of ``false alarms", where the formal model indicates a behaviour as unsafe, although it turns out to be uncritical when analysed in simulation.
In scenario A, both abstraction levels were relatively consistent, with a total of 22 behaviours being identified as unsafe by both formal model and simulation. There was one instance of a missed alarm and no false alarms. In the comparatively more complex scenario B, the disagreements between abstraction layers were more pronounced, with 44 agreements, 11 missed alarms and 39 false alarms. Closer investigation of the results showed that both missed and false alarms are consequences of modelling abstractions in the formal models. For example, in scenario B, the simulation-based MCTS search found hazardous behaviours in which the worker activates the robot by accidentally pressing an activation button with his elbow while still bent over the workpiece, resulting in a collision between robot and human head. Effects like these were not considered in sufficient detail in our formal model, thus leading to missed alarms. False alarms appeared mainly due to coarse workspace discretization, which led the formal model to indicate collisions that were not confirmed when analysing the same sequence in the simulator.  

\section{DISCUSSION AND FUTURE WORK}
\label{sec:discussion}
We have demonstrated a novel hazard analysis approach which relies on analyses in two distinct modelling layers. The higher abstraction level of the first layer is used as preprocessing step for the lower-layer analysis. For that, we propose to leverage supervisory control theory to integrate the two layers. The main value of this two-layer approach is to identify a set of \textit{potentially} critical cases which are then closer examined in simulation. Deploying the formal model as a first analysis step leverages system knowledge to increase search efficiency.

In our experiments, the two-layer approach found more hazards than approaches based on simulation only. The experiments also highlight the dangers of missed and false alarms. Especially missed alarms are critical, as they result in hazards remaining undetected. This can be addressed by improving the formal models. However, in doing so, users must take care since introducing additional model components may restrict the overall system behaviour in an unintended way (e.g., a guard statement in one component of the model may inadvertently block an event in another model). Alternatively, users may consider to formulate safety criteria that are even more conservative, so that the formal model is biased towards false alarms, thereby over-approximating the set of potentially unsafe behaviours. This leads to more computational effort, as more potentially hazardous sequences are passed to the simulation, but is likely to decrease the number of missed alarms.
It may also be worthwhile to split the computational budget between a two-layer approach and a simulation-only search (e.g., MCTS). With the latter being not dependant on a formal model, the overall analysis would be more robust against missed alarms of the formal model.
Yet, we emphasise that even when considering the missed alarms, the two-layer approach still outperformed its baselines. Also, the missed alarms in our experiments did not reflect principal insufficiency of the proposed synthesis, but rather the simplicity of the formal models that we chose as a basis for our experiments.
Finally, it should be noted that while this paper is focused on comparing formal model and simulation model, similar problems can also arise when comparing simulation model and real world system. After all, any tool or method used for hazard analysis contains abstractions users need to be aware of.

In future work, the influence of varying abstraction levels on the hazard analysis needs to be investigated in a more comprehensive manner. We will also consider introducing postprocessing methods, such as clustering the unsafe sequences, as multiple sequences may just be variations of the same hazard manifesting in different ways.
Another interesting aspect to consider is the possibility of learning automata models from simulation \cite{selvaraj2022automatically,farooqui2018towards}. This will introduce a feedback loop whereby the abstract layer first generates inputs for the simulation layer, but is then refined by learning from the executions in simulation.

\section*{ACKNOWLEDGMENT}
The authors thank Tamim Asfour for his support.


\IEEEtriggeratref{18}
\bibliography{references.bib}
\bibliographystyle{IEEEtran}
\end{document}

%% file: methodA_YS.tex
\newcommand{\SUS}{collaborative system\xspace}

To formally analyse unsafe behaviours, we model the \SUS as a DES, which is a discrete-state, event-driven system that occupies at each time instance a single \emph{state} out of its many possible ones, and transits to another state on the occurrence of an \emph{event}. Thus, a characteristic feature of DES is the notion of instantaneous events that may be associated with common phenomena in a \SUS, such as a collision or a human pressing a button. DES can be modelled and analysed in varying levels of detail~\cite{CasLaf:08}.

As a DES modelling formalism, we use Extended Finite Automata (EFAs)~\cite{skoldstam2007modeling}. An EFA extends finite automata with bounded discrete variables, guards (logical expressions) over the variables, and actions that assign values to the variables on the transitions. 

\begin{definition}
An EFA is a tuple
\begin{equation*}
    E = \langle \Sigma, V, L, \to, L^i, L^m\rangle
\end{equation*}
where $\Sigma$ is a finite set of events, $V = \langle v_1, v_2,\dots, v_n\rangle$ is a finite set of bounded discrete variables, $L$ is a finite set of locations, $\to\,\subseteq L \times \Sigma \times G \times A \times L$ is the conditional transition relation, where $G$ and $A$ are the respective sets of guards and actions, $L^i \subseteq L$ is the set of initial locations, and $L^m \subseteq L$ is the set of marked locations.
\end{definition}

The current state of an EFA is given by its current location together with the current values of the variables. Let each variable $v_i \in V$ be associated with a bounded discrete domain $\hat{v}_i$ and $\domv = \hat{v}_1\times \hat{v}_2\times\dots\times \hat{v}_n$ be the domain of $V$. The set of states of an EFA is given by $Q = L\times \domv$. The expression $l_0 \xrightarrow{\sigma:[g]a} l_1$ denotes a transition from location $l_0$ to $l_1$ labelled by event $\sigma \in \Sigma$, and with guard $g \in G$ and action $a\in A$. The transition is enabled when $g$ evaluates to \emph{true}. On occurrence of $\sigma$, $a$ updates some of the values of the variables $v \in V$, thereby causing the EFA to change location from $l_0$ to $l_1$. Note that \EFAs possess the same expressive power as finite automata (FA) and can be transformed into equivalent FA~\cite{skoldstam2007modeling}. However, the richer structure of \EFAs provide more compact models compared to FAs.

\EFAs naturally interact through shared variables, but can also interact through shared events, which is modelled by \emph{synchronous composition}. Common events occur simultaneously in all interacting \EFAs, while local, non-shared events occur independently.
$E_1\parallel E_2$ denotes the synchronous composition of the \EFAs $E_1$ and $E_2$.
As defined by~\cite{skoldstam2007modeling}, transitions that are labelled by shared events but have mutually exclusive guards, or transitions that have conflicting actions, can never occur. This interaction mechanism provides an efficient way to model complex
systems as a set of interacting \EFAs in a modular way and to compositionally reason about their overall behaviour. 

A DES model of the \SUS is an abstraction of the feasible interactions between the subsystems, including those that lead to unsafe states. Ideally, the hazard analysis on this model should result in finding all behaviours that lead to unsafe states. In this paper, we use synthesis algorithms from SCT to identify such unsafe behaviours. SCT provides a framework for modelling, synthesis, and verification of reactive control functions for DES~\cite{ramadge1989control}. Given a DES model of a system to control, the \emph{plant} $G$, and a \emph{specification} $K$ of the desired controlled behaviour, SCT provides means to synthesise a \emph{supervisor} that, interacting with the plant in a \emph{closed-loop}, dynamically restricts the event generation of the plant such that the specification is satisfied.

Supervisor synthesis is an iterative fixpoint algorithm that removes some behaviour from the synchronous composition $G\parallel K$ such that the resulting supervisor, and hence the closed-loop system, is guaranteed to fulfil certain properties, in addition to the specification~\cite{ramadge1989control,CasLaf:08}. Two such properties relevant in our context are \emph{non-blocking} and \emph{minimally-restrictive}. The non-blocking property requires the supervisor to guarantee that the system can reach some marked state from any reachable state; and the minimally restrictive property requires the supervisor to minimise the behaviour that is removed. An  illustration of the synthesis algorithm is given in Fig.~\ref{fig:synth:example}, where the plant $G$ and specification $K$ are shown in Fig.~\ref{fig:G} and Fig.~\ref{fig:K}, respectively. The minimally-restrictive supervisor synthesised from $G\parallel K$ only removes those states and transitions that break the non-blocking property as shown in Fig.~\ref{fig:GsyncK}. The software tool Supremica~\cite{supremica2017} implements such synthesis algorithms and other techniques for modelling and analysis of DES. 

For a supervisor synthesis problem, the specification typically describes \emph{desired} behaviour through marked states in the DES model. Although it may seem counter-intuitive, we use marked states to specify \emph{undesired}, i.e. hazardous, states of the \SUS. Through synthesising a minimally-restrictive non-blocking supervisor, the largest subset of behaviours that lead to hazardous system states can then be obtained. In this approach, marked states, such as human-robot collision states, may be specified based on human reasoning and expert knowledge, complemented by established hazard analysis checklists. Accordingly, such specification of marked states may inform what a suitable level of abstraction in the DES modelling of the \SUS would be. The algorithmically generated set of corresponding hazardous behaviours can then be further analysed in simulation.

\begin{figure}[t]
    \centering
    \subfloat[][$G$]{
        \centering
        \begin{tikzpicture} [every node/.style={scale=0.5},node distance = 2cm, on grid,auto]

        
        \node (q0) [state,
        	initial,
        	initial left,
        	initial distance=0.25cm,
        	initial text=$G$:,
        ] {\texttt{i}};
        \node (q1) [state, below = of q0, accepting] {\texttt{j}};
        \node (q2) [state, right = of q0] {\texttt{k}};

        \node[rectangle] at (0, -2.48)  (inv)     {};
        
        \path [-stealth, thick]
        (q0) edge [bend left] node {a}   (q1)
        (q0) edge [bend left] node {c}   (q2)
        (q2) edge [bend left] node {d}   (q0)
        (q1) edge [bend left] node {b}   (q0);
        
        \end{tikzpicture}
        \label{fig:G}
    }
    \subfloat[][$K$]{
        \centering
        \begin{tikzpicture} [every node/.style={scale=0.5},node distance = 2cm, on grid,auto]

        
        \node (q0) [state,
        	initial,
        	initial left,
        	initial distance=0.25cm,
        	initial text=$K$:,
        ] {\texttt{x}};
        \node (q1) [state, below = of q0, accepting] {\texttt{y}};
        
        \path [-stealth, thick,  every loop/.style={min distance=5mm,looseness=1}]
        (q0) edge [] node {a}   (q1)
        (q0) edge[loop] node [above]  {b, c, d} (q0)
        (q1) edge[loop below ] node {b, c, d} (q1);
        
        \end{tikzpicture} 
        \label{fig:K}
    }
    \subfloat[][$G\parallel K$]{
        \centering
        \begin{tikzpicture} [every node/.style={scale=0.5},node distance = 2cm, on grid,auto]

        
        \node (q0) [state,
        	initial,
        	initial left,
        	initial distance=0.25cm,
        	initial text=$G\parallel K$:,
        ] {$\langle\texttt{i,x}\rangle$};
        \node (q1) [draw=black!25,state, right = of q0,text=black!25] {$\langle\texttt{k,y}\rangle$};
        \node (q2) [state, below = of q0] {$\langle \texttt{k,x}\rangle$};
        \node (q3) [draw=black!25,state, right = of q2,text=black!25] {$\langle\texttt{i,y}\rangle$};
        \node (qm) [draw=blue,state, below right = 1.414cm of q0, text=blue,accepting] {$\langle\texttt{j,y}\rangle$};        
        \node[rectangle] at (0, -2.55)  (inv)     {};
        
        \path [-stealth, thick]
        (q0) edge [] node[text=black] {a}   (qm)
        (qm) edge [draw=black!25] node[text=black!25,below] {b}   (q3)
        
        
        
        (q0) edge [draw=black,bend left] node[text=black] {c}   (q2)
        (q2) edge [draw=black,bend left] node[text=black] {d}   (q0)
        
        (q1) edge [draw=black!25,bend left] node[text=black!25] {d}   (q3)
        (q3) edge [draw=black!25,bend left] node[text=black!25] {c}   (q1);
        
        \end{tikzpicture}
        \label{fig:GsyncK}
    }
    \caption{Illustration of supervisor synthesis. The marked states are indicated by double circles. The grey states and transitions in $G\parallel K$ are disabled by the minimally-restrictive non-blocking supervisor.}
    \label{fig:synth:example}
\end{figure}

%% file: automatonFig_combined.tex
\begin{figure*}[ht!]
		\centering
		\resizebox{0.47\textwidth}{!}{
		\begin{tikzpicture}[every node/.style=->,>=stealth',node distance = 5cm,thick,on grid,auto]
			\node (qH0) [state, initial, initial above, initial text ={$\mathcal{H}$:}] {$q_{\mathcal{H}0}$};
            \node (qH1) [state, left =3cm of qH0]{$q_{\mathcal{H}1}$};
            \node (qH2) [state, below = of qH1]{$q_{\mathcal{H}2}$};
            \node (qH3) [state, below = of qH0]{$q_{\mathcal{H}3}$};
            \node (qH4) [state, right =4cm of qH0]{$q_{\mathcal{H}4}$};
            \node (qH5) [state, below = of qH4]{$q_{\mathcal{H}5}$};

			\path[->]
            (qH0) edge [bend right=40] node[above] {$t_1$} (qH1)
			(qH1) edge node[above] {$t_1$} (qH0)
			
			(qH1) edge [bend left] node [above, rotate=-90, xshift=0.5cm, text width=3.5cm] {$\color{black}u_S:\color{blue}[P=0] \ \  \color{red} P\leftarrow1\color{black}$} (qH2)
			(qH2) edge [bend left] node [below, rotate=-90, xshift=0.5cm, text width=3.5cm] {$\color{black}d_S:\ \color{blue} [P=1]\ \color{red} P\leftarrow0\color{black}$} (qH1)
			(qH2) edge node [below, rotate=-90, xshift=0cm] {$r$} (qH1)
			
			(qH0) edge [bend left] node [above, rotate=-90, xshift=0.5cm, text width=3cm] {$b_1,b_2,b_3:\ \color{blue} [P\neq 1]\ \color{black} $} (qH3)
			(qH3) edge [bend left] node [below, rotate=-90] {$r$} (qH0)
			
			(qH0) edge [bend left=40] node [above, text width=2cm, xshift=0.2cm] {$t_2\ \color{red} S\leftarrow1$\color{black}} (qH4)
			(qH4) edge node [above, text width=2cm, xshift=0.2cm] {$t_2\ \color{red} S\leftarrow0\color{black}$} (qH0)
			
			(qH4) edge [bend left] node [above, rotate=-90, xshift=0cm, text width=5cm] {$u_R:\ \color{blue}[P=2]\ \color{red}\ P\leftarrow1,\ W\leftarrow 1\color{black} $} (qH5)
			(qH5) edge node [above, rotate=-90, xshift=0.3cm, text width=2cm] {$r\  \color{red}W\leftarrow 0\color{black}$} (qH4)
			(qH4) edge [bend right] node [below, rotate=-90, xshift=0.3cm, text width=5cm] {$d_R:\ \color{blue}[P=1]\ \color{red}P\leftarrow2,\ W\leftarrow 1\color{black} $} (qH5);
		\end{tikzpicture}}
		\resizebox{0.43\textwidth}{!}{
		\begin{tikzpicture}[every node/.style=->,>=stealth',node distance = 6cm,thick,on grid,auto]
        	\node [state, initial, initial above,initial text ={$\mathcal{R}$:}] (qR0) at (4.5,-4)     {$q_{R0}$};
			\node [state] (qR2) at (0,-4)     {$q_{\mathcal{R}2}$};
			\node [state] (qR1) at (4.5,-9)     {$q_{\mathcal{R}1}$};

			\path[->]
			(qR0) edge [bend left=0] node {$b_2:\ \color{blue}[P\neq1]\ \color{red}\ R\leftarrow1\color{black}$} (qR2)
			(qR2) edge [bend left=40] node {$\color{black}b_0:\ \color{blue}[P\neq1]\ \color{red}\ R\leftarrow0\color{black}$} (qR0)
			
			(qR2) edge [bend right=20] node [above, rotate=-50, xshift=0cm, text width=3cm] {$\color{black}\ b_1:\ \color{blue}[P\neq1]$} (qR1)
			(qR1) edge [bend left=60] node {$\color{black}\ b_2:\ \color{blue}[P\neq1]$} (qR2)
			
			(qR0) edge [bend right] node [below, rotate=-90, xshift=0cm, text width=3.5cm] {$\color{black}\ b_1:\ \color{blue}[P\neq1]\ \color{red}R\leftarrow1\color{black}$} (qR1)
			(qR1) edge [bend right=0] node [above, rotate=-90, xshift=0cm, text width=3.5cm] {$\color{black}\ b_0:\ \color{blue}[P\neq1]\ \color{red}R\leftarrow0\color{black}$} (qR0)
			(qR1) edge [bend right=40] node [above, rotate=-90, xshift=1cm, text width=6cm] {$\color{black}\ safetyStop:\ \color{blue}[S=1]:\ \color{red}R\leftarrow0\color{black}$} (qR0)
			
			(qR0) edge [loop right] node {$\color{black}b_0:\ \color{blue}[P\neq1]$} ()
			
			(qR1) edge [loop below] node {$\color{black}b_1: \color{blue}[P\neq1]$} ()
			
			(qR2) edge [loop left] node {$\color{black}b_2:\ \color{blue}[P\neq1]\ $} ();
		\end{tikzpicture}}
		\resizebox{0.05\textwidth}{!}{
    		\begin{tikzpicture}[every node/.style=->,>=stealth',node distance = 5cm,thick,on grid,auto]
    		\node (qS1) [state, initial, initial above, initial text ={$\mathcal{SP}$:}]  {$q_{\mathcal{SP}0}$};
    		\node (qS2) [state,below=of qS1, accepting] {$q_{\mathcal{SP}1}$};

    		\path[->]
            (qS1) edge  node [above, rotate=-90, xshift=1cm, text width=6cm] {\begin{tabular}{c} $\color{black} c:\ \color{blue}[R=1 \land W=1]$ \end{tabular}} (qS2);          
    	\end{tikzpicture}}
		\caption{EFA models of human worker $\mathcal{H}$ (left), robot $\mathcal{R}$ (middle), and safety specification $\mathcal{SP}$ (right). Guards are denoted in blue, actions in red. Marked locations are denoted by double circles. A transition can only be taken if the guard expression evaluates to \textit{true}. When the transition is taken, variables are updated according to the action.}
		\label{fig:example}
	\end{figure*}

%% file: root.bbl
\begin{thebibliography}{10}
\providecommand{\url}[1]{#1}
\csname url@samestyle\endcsname
\providecommand{\newblock}{\relax}
\providecommand{\bibinfo}[2]{#2}
\providecommand{\BIBentrySTDinterwordspacing}{\spaceskip=0pt\relax}
\providecommand{\BIBentryALTinterwordstretchfactor}{4}
\providecommand{\BIBentryALTinterwordspacing}{\spaceskip=\fontdimen2\font plus
\BIBentryALTinterwordstretchfactor\fontdimen3\font minus
  \fontdimen4\font\relax}
\providecommand{\BIBforeignlanguage}[2]{{%
\expandafter\ifx\csname l@#1\endcsname\relax
\typeout{** WARNING: IEEEtran.bst: No hyphenation pattern has been}%
\typeout{** loaded for the language `#1'. Using the pattern for}%
\typeout{** the default language instead.}%
\else
\language=\csname l@#1\endcsname
\fi
#2}}
\providecommand{\BIBdecl}{\relax}
\BIBdecl

\bibitem{STD_ISO2011}
{ISO}, ``{ISO 12100:2011}: {Safety of machinery - General principles for design
  - Risk assessment and risk reduction},'' 2011.

\bibitem{STD_IEC61508}
{IEC}, ``{IEC} 61508-1:2010-1 {F}unctional safety of
  electrical/electronic/programmable electronic safety-related systems - {P}art
  1: {G}eneral requirements,'' {I}nternational {E}lectrotechnical {C}ommission,
  2006.

\bibitem{Hornung2021}
L.~Hornung and C.~Wurll, ``Human-robot collaboration: a survey on the state of
  the art focusing on risk assessment,'' in \emph{Berichte aus der Robotik -
  Robotix-Academy Conference for Industrial Robotics (RACIR) 2021}, Sep. 2021,
  pp. 10--17.

\bibitem{RA_Leveson2012}
N.~Leveson, \emph{Engineering a safer world: Systems thinking applied to
  safety}.\hskip 1em plus 0.5em minus 0.4em\relax MIT {P}ress, 2011.

\bibitem{Askarpour2016}
M.~Askarpour, D.~Mandrioli, M.~Rossi, and F.~Vicentini, ``Safer-hrc: Safety
  analysis through formal verification in human-robot collaboration,'' in
  \emph{International Conference on Computer Safety, Reliability, and
  Security}.\hskip 1em plus 0.5em minus 0.4em\relax Springer, 2016, pp.
  283--295.

\bibitem{RA_Askarpour2021}
M.~Askarpour, L.~Lestingi, S.~Longoni, N.~Iannacci, M.~Rossi, and F.~Vicentini,
  ``Formally-based model-driven development of collaborative robotic
  applications,'' \emph{Journal of Intelligent \& Robotic Systems}, vol. 102,
  no.~3, 2021.

\bibitem{RA_Rathmair2021}
M.~Rathmair, C.~Luckeneder, T.~Haspl, B.~Reiterer, R.~Hoch, M.~Hofbaur, and
  H.~Kaindl, ``Formal verification of safety properties of collaborative
  robotic applications including variability,'' in \emph{2021 30th IEEE
  International Conference on Robot \& Human Interactive Communication
  (RO-MAN)}.\hskip 1em plus 0.5em minus 0.4em\relax IEEE, 2021, pp. 1283--1288.

\bibitem{RA_Araiza2016}
D.~Araiza-Illan, A.~G. Pipe, and K.~Eder, ``Intelligent agent-based stimulation
  for testing robotic software in human-robot interactions,'' in
  \emph{Proceedings of the 3rd Workshop on Model-Driven Robot Software
  Engineering}, 2016, pp. 9--16.

\bibitem{RA_Bobka2016a}
P.~Bobka, T.~Germann, J.~K. Heyn, R.~Gerbers, F.~Dietrich, and K.~Dröder,
  ``Simulation platform to investigate safe operation of human-robot
  collaboration systems,'' in \emph{6th CIRP Conference on Assembly
  Technologies and Systems (CATS)}, vol.~44, 2016, pp. 187 -- 192.

\bibitem{RA_Huck2021}
T.~Huck, C.~Ledermann, and T.~Kröger, ``Virtual adversarial humans finding
  hazards in robot workplaces,'' in \emph{2021 IEEE International Conference on
  Robotics and Automation (ICRA)}.\hskip 1em plus 0.5em minus 0.4em\relax IEEE,
  2021.

\bibitem{Dill1998}
D.~L. Dill, ``What's between simulation and formal verification?'' in
  \emph{Proceedings 1998 Design and Automation Conference. 35th DAC.(Cat. No.
  98CH36175)}.\hskip 1em plus 0.5em minus 0.4em\relax IEEE, 1998, pp. 328--329.

\bibitem{ramadge1989control}
P.~J. Ramadge and W.~M. Wonham, ``The control of discrete event systems,''
  \emph{Proceedings of the IEEE}, vol.~77, no.~1, pp. 81--98, 1989.

\bibitem{STD_IEC61882}
``{IEC} 61882:2016: {H}azard and operability studies ({HAZOP} studies) -
  application guide,'' {I}nternational {E}lectrotechnical {C}ommission, 2016.

\bibitem{Guiochet2016}
J.~Guiochet, ``Hazard analysis of human--robot interactions with hazop--uml,''
  \emph{Safety science}, vol.~84, pp. 225--237, 2016.

\bibitem{RA_Awad2017}
R.~{Awad}, M.~{Fechter}, and J.~{van Heerden}, ``{Integrated risk assessment
  and safety consideration during design of HRC workplaces},'' in \emph{22nd
  IEEE International Conference on Emerging Technologies and Factory Automation
  (ETFA)}, Sep. 2017.

\bibitem{BaiKat:08}
C.~Baier and J.-P. Katoen, \emph{Principles of Model Checking}.\hskip 1em plus
  0.5em minus 0.4em\relax {MIT} Press, 2008.

\bibitem{MISC_Clarke2018}
E.~M. Clarke, T.~A. Henzinger, H.~Veith, R.~Bloem \emph{et~al.}, \emph{Handbook
  of model checking}.\hskip 1em plus 0.5em minus 0.4em\relax Springer, 2018,
  vol.~10.

\bibitem{Askarpour2017}
M.~Askarpour, D.~Mandrioli, M.~Rossi, and F.~Vicentini, ``Modeling operator
  behavior in the safety analysis of collaborative robotic applications,'' in
  \emph{International Conference on Computer Safety, Reliability, and
  Security}.\hskip 1em plus 0.5em minus 0.4em\relax Springer, 2017, pp.
  89--104.

\bibitem{RA_Askarpour2020}
M.~Askarpour, M.~Rossi, and O.~Tiryakiler, ``{Co-simulation of human-robot
  collaboration: From temporal logic to 3D simulation},'' in \emph{1st Workshop
  on Agents and Robots for Reliable Engineered Autonomy, AREA 2020}, vol.
  319.\hskip 1em plus 0.5em minus 0.4em\relax Open Publishing Association,
  2020, pp. 1--8.

\bibitem{Corso2020Survey}
A.~Corso, R.~Moss, M.~Koren, R.~Lee, and M.~Kochenderfer, ``A survey of
  algorithms for black-box safety validation of cyber-physical systems,''
  \emph{Journal of Artificial Intelligence Research}, vol.~72, 2021.

\bibitem{Norden2019}
J.~Norden, M.~O'Kelly, and A.~Sinha, ``Efficient black-box assessment of
  autonomous vehicle safety,'' \emph{arXiv preprint arXiv:1912.03618}, 2019.

\bibitem{Ding2020}
W.~Ding, B.~Chen, M.~Xu, and D.~Zhao, ``Learning to collide: An adaptive
  safety-critical scenarios generating method,'' in \emph{2020 IEEE/RSJ
  International Conference on Intelligent Robots and Systems (IROS)}.\hskip 1em
  plus 0.5em minus 0.4em\relax IEEE, 2020.

\bibitem{Chance2019}
G.~Chance, A.~Ghobrial, S.~Lemaignan, T.~Pipe, and K.~Eder, ``An
  agency-directed approach to test generation for simulation-based autonomous
  vehicle verification,'' \emph{arXiv preprint arXiv:1912.05434}, 2019.

\bibitem{Lee2015}
R.~Lee, M.~J. Kochenderfer, O.~J. Mengshoel, G.~P. Brat, and M.~P. Owen,
  ``Adaptive stress testing of airborne collision avoidance systems,'' in
  \emph{2015 IEEE/AIAA 34th Digital Avionics Systems Conference (DASC)}.\hskip
  1em plus 0.5em minus 0.4em\relax IEEE, 2015.

\bibitem{Webster2020}
M.~Webster, D.~Western, D.~Araiza-Illan, C.~Dixon, K.~Eder, M.~Fisher, and
  A.~G. Pipe, ``A corroborative approach to verification and validation of
  human--robot teams,'' \emph{The International Journal of Robotics Research},
  vol.~39, no.~1, pp. 73--99, 2020.

\bibitem{Askarpour2020}
M.~Askarpour, M.~Rossi, and O.~Tiryakiler, ``{Co-simulation of human-robot
  collaboration: From temporal logic to 3D simulation},'' in \emph{1st Workshop
  on Agents and Robots for Reliable Engineered Autonomy, AREA 2020}, vol.
  319.\hskip 1em plus 0.5em minus 0.4em\relax Open Publishing Association,
  2020, pp. 1--8.

\bibitem{CasLaf:08}
C.~G. Cassandras and S.~Lafortune, \emph{Introduction to Discrete Event
  Systems}, 2nd~ed.\hskip 1em plus 0.5em minus 0.4em\relax New York, NY, USA:
  Springer Science \& Business Media, 2008.

\bibitem{skoldstam2007modeling}
M.~Skoldstam, K.~Akesson, and M.~Fabian, ``Modeling of discrete event systems
  using finite automata with variables,'' in \emph{2007 46th IEEE Conference on
  Decision and Control}.\hskip 1em plus 0.5em minus 0.4em\relax IEEE, 2007, pp.
  3387--3392.

\bibitem{supremica2017}
R.~Malik, K.~Akesson, H.~Flordal, and M.~Fabian, ``Supremica-{A}n efficient
  tool for large-scale discrete event systems,'' \emph{IFAC-PapersOnLine},
  vol.~50, no.~1, pp. 5794 -- 5799, 2017, 20th IFAC World Congress.

\bibitem{STD_ISOTS15066}
``{ISO} {TS} 15066:2016 {R}obots and robotic devices - {C}ollaborative
  robots,'' {I}nternational {O}rganization for {S}tandardization, 2016.

\bibitem{CoppeliaSim2013}
E.~Rohmer, S.~P.~N. Singh, and M.~Freese, ``Coppeliasim (formerly v-rep): a
  versatile and scalable robot simulation framework,'' in \emph{Proc. of The
  International Conference on Intelligent Robots and Systems (IROS)}, 2013,
  www.coppeliarobotics.com.

\bibitem{selvaraj2022automatically}
Y.~Selvaraj, A.~Farooqui, G.~Panahandeh, W.~Ahrendt, and M.~Fabian,
  ``Automatically learning formal models from autonomous driving software,''
  \emph{Electronics}, vol.~11, no.~4, p. 643, 2022.

\bibitem{farooqui2018towards}
A.~Farooqui, P.~Falkman, and M.~Fabian, ``Towards automatic learning of
  discrete-event models from simulations,'' in \emph{2018 IEEE 14th
  International Conference on Automation Science and Engineering (CASE)}.\hskip
  1em plus 0.5em minus 0.4em\relax IEEE, 2018, pp. 857--862.

\end{thebibliography}
